\documentclass[final]{cvpr}

\usepackage{times}
\usepackage{epsfig}
\usepackage{graphicx}
\usepackage{amsmath}
\usepackage{amssymb}

\usepackage[nocompress]{cite}
\usepackage{booktabs}
\usepackage{caption}
\usepackage{placeins}
\usepackage{mwe}

\usepackage[pagebackref=true,breaklinks=true,colorlinks,bookmarks=false]{hyperref}

\def\confYear{CVPR 2021}

\begin{document}

\title{Multi-View Multi-Person 3D Pose Estimation with Plane Sweep Stereo}

\author{Jiahao Lin {} {} {} {} {} {} {} {} Gim Hee Lee\\
Department of Computer Science, National University of Singapore\\
{\tt\small \{jiahao, gimhee.lee\}@comp.nus.edu.sg}
}

\maketitle

\begin{abstract}

Existing approaches for multi-view multi-person 3D pose estimation explicitly establish cross-view correspondences to group 2D pose detections from multiple camera views and solve for the 3D pose estimation for each person.
Establishing cross-view correspondences is challenging in multi-person scenes, and incorrect correspondences will lead to sub-optimal performance for the multi-stage pipeline.
In this work, we present our multi-view 3D pose estimation approach based on plane sweep stereo to jointly address the cross-view fusion and 3D pose reconstruction in a single shot.
Specifically, we propose to perform depth regression for each joint of each 2D pose in a target camera view.
Cross-view consistency constraints are implicitly enforced by multiple reference camera views via the plane sweep algorithm to facilitate accurate depth regression.
We adopt a coarse-to-fine scheme to first regress the person-level depth followed by a per-person joint-level relative depth estimation.
3D poses are obtained from a simple back-projection given the estimated depths.
We evaluate our approach on benchmark datasets where it outperforms previous state-of-the-arts while being remarkably efficient.
Our code is available at the project website. \footnote{\url{https://github.com/jiahaoLjh/PlaneSweepPose}}

\end{abstract}

\section{Introduction}

\FloatBarrier
\begin{figure}[t]
    \centering
    \includegraphics[width=\columnwidth, trim=30 130 120 20, clip]{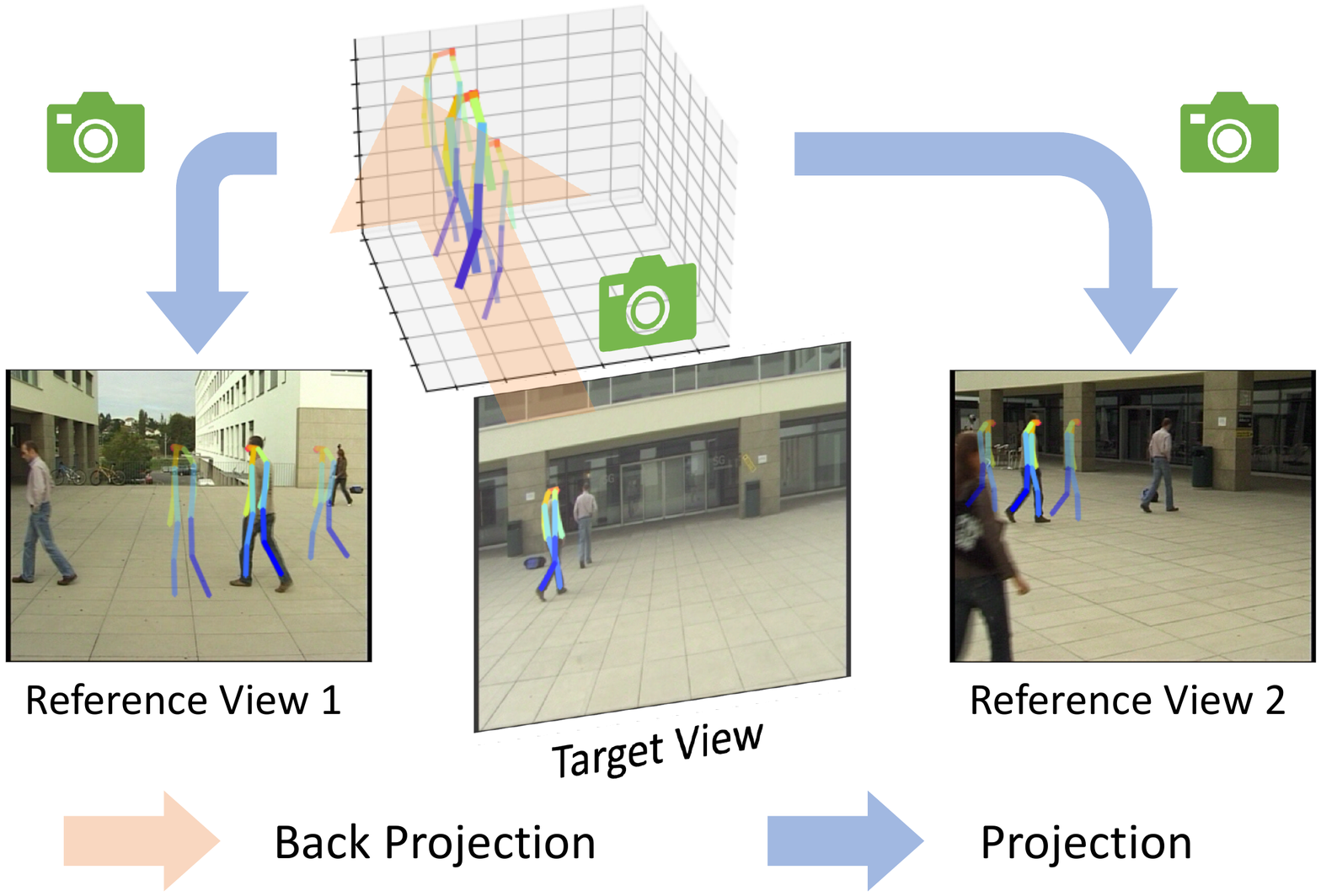}
    \caption{Our method is based on plane sweep stereo to regress depths for 2D pose detections. 2D poses are back-projected to successive depth planes and warped to reference views for consistency measurement which is utilized for depth regression.}
    \label{fig:teaser}
    \vspace{-1mm}
\end{figure}

3D human pose estimation has been an active research area in the field of computer vision due to its large number of real-world applications such as human-computer interaction, virtual and augmented reality, camera surveillance, \textit{etc}. However, 3D human pose estimation for multiple persons from monocular images is an ill-posed and challenging problem due to both the loss of depth information and severe occlusions under a single camera viewpoint.
On the other hand, multi-view images captured by multiple cameras provide complementary information of the scene that can be used to effectively alleviate projective
ambiguities.

Unlike its multi-view single person \cite{qiu2019cross,iskakov2019learnable,remelli2020lightweight,iqbal2020weakly,xie2020metafuse} counterpart, the fusion of information from multi-view images with multiple persons is more challenging since the identity of the 2D poses from each camera view is unknown.
Previous works such as \cite{dong2019fast,chen2020multi,huang2020end} address this problem in three steps. The 2D poses are first estimated for each camera view independently.
Subsequently, the 2D poses from different views that correspond to the same person are identified and grouped together.
Finally, the 3D pose of each person is estimated  
with triangulation or optimization-based pictorial structure models
using the set of grouped 2D pose detections from multiple views. 

The establishment of cross-view correspondences is critical for multi-view multi-person 3D pose estimation.
Traditional methods use either greedy matching approach \cite{huang2020end} for fast inference speed, or optimization-based approach \cite{dong2019fast,ershadi2018multiple,chen2020multi} for better global consistency.
Recently, VoxelPose \cite{tu2020voxelpose} is proposed to jointly solve the challenging cross-view matching and 3D pose estimation problems in an object-detection paradigm.
Instead of explicitly searching for 2D pose correspondences, VoxelPose projects the 2D pose heatmaps from multiple views to a common 3D space, and performs both 3D pose detection and estimation in the 3D volumetric space.
The 3D object detection formulation avoids the explicit cross-view matching step, thus effectively reduces the impact from incorrectly established cross-view correspondences.
Despite its effectiveness, several limitations exist for the object-detection-based pipeline:
1) Prior knowledge of the common 3D space dimension according to the multi-camera settings is needed to define the volumetric space for 3D object detection.
2) The back-projection of the 2D pose detections on
each 3D voxel is not scalable to larger scenes.
3) 3D convolution that is applied to all voxel locations incurs unnecessary heavy computations, especially for large sparse scenes.

In this work, we present our plane-sweep-based approach for multi-view multi-person 3D pose estimation.
Our approach avoids explicit cross-view matching and aggregates multiple views for 3D pose estimation in a single shot.
Specifically, we build our framework upon the concept of plane sweep stereo \cite{collins1996space} to estimate the depth for each joint of each person in a target camera view.
As illustrated in Figure \ref{fig:teaser}, 2D poses are first back-projected to successive virtual depth planes, and then warped to the respective reference camera views.
We measure the cross-view consistency at each depth level,
which is then used to regress the depths from standard convolutional neural networks.
Our depth regression adopts a two-stage coarse-to-fine scheme.
Person-level depth is first estimated for each 2D pose.
Joint-level relative depth with respect to the person-level depth within a much smaller depth range is then regressed for each joint.
The two stages can be trained together in an end-to-end manner.
During inference, we obtain the 3D poses by back-projecting the 2D poses with the estimated depths.
Multiple 3D poses of the same person from different views can be easily merged via a simple distance-based clustering.

We evaluate our plane-sweep-based framework on three benchmark datasets, \textit{i.e.}, the Campus and the Shelf datasets, and CMU Panoptic dataset, where we outperform existing state-of-the-arts.
In addition to the removal of explicit cross-view matching and triangulation compared to the traditional three-step approaches, our method is also more efficient compared to VoxelPose in two aspects: 1) In contrast to VoxelPose that builds voxels in the 3D space, we leverage on the plane sweep algorithm
that is proportional to only the number of virtual depth planes.
2) Instead of performing 3D convolution on all voxel locations, we
utilize the much faster 1D convolutions for each 2D pose.
Furthermore, our method is more generablizable to scenarios with no prior knowledge of the multi-camera settings since only the range of virtual depth planes needs to be pre-defined for each camera view.

Our contributions in this work are:
\begin{itemize}
    \item We present a plane-sweep-based approach to perform multi-view multi-person 3D pose estimation without the need for explicit cross-view matching.
    \item Our approach outperforms existing state-of-the-arts on benchmark datasets, while being much more efficient compared to existing works.
\end{itemize}

\section{Related Work}

In this section, we briefly review the related works that utilize multiple camera views for 3D pose estimation.

\subsection{Multi-view Single-person}
3D human pose estimation from 2D images is an ill-posed problem due to the loss of depth information in the process of camera projection.
Exploiting multi-view images is an effective way to alleviate projective ambiguities since multiple camera viewpoints provide complementary information of the 3D scene.
Extensive research has been done for the single-person 3D pose estimation task under the multi-view setting.
Qiu \textit{et al.} \cite{qiu2019cross} propose to fuse multiple views in the feature space with the epipolar geometry \cite{hartley2003multiple} for more accurate 2D pose estimates. A recursive pictorial structure model is used to reconstruct the 3D pose from the multi-view 2D detections.
The idea of explicit fusion is also adopted in \cite{remelli2020lightweight} by Remelli \textit{et al.} They transform the latent features with the known camera extrinsics into a canonical 3D space, where the transformed features from multiple views are then stacked together for joint reasoning.
Iskakov \textit{et al.} \cite{iskakov2019learnable} present a learnable triangulation method that learns per-view confidence weights for the standard triangulation \cite{hartley2003multiple}, and a volumetric-based method that aggregates multi-view images and performs the 3D pose estimation in a 3D volumetric space.
Weakly-supervised approach \cite{iqbal2020weakly} and meta-learning approach \cite{xie2020metafuse} have also been proposed to utilize the multi-view settings.

\begin{figure*}[t]
    \centering
    \includegraphics[width=\linewidth, trim=30 135 50 65, clip]{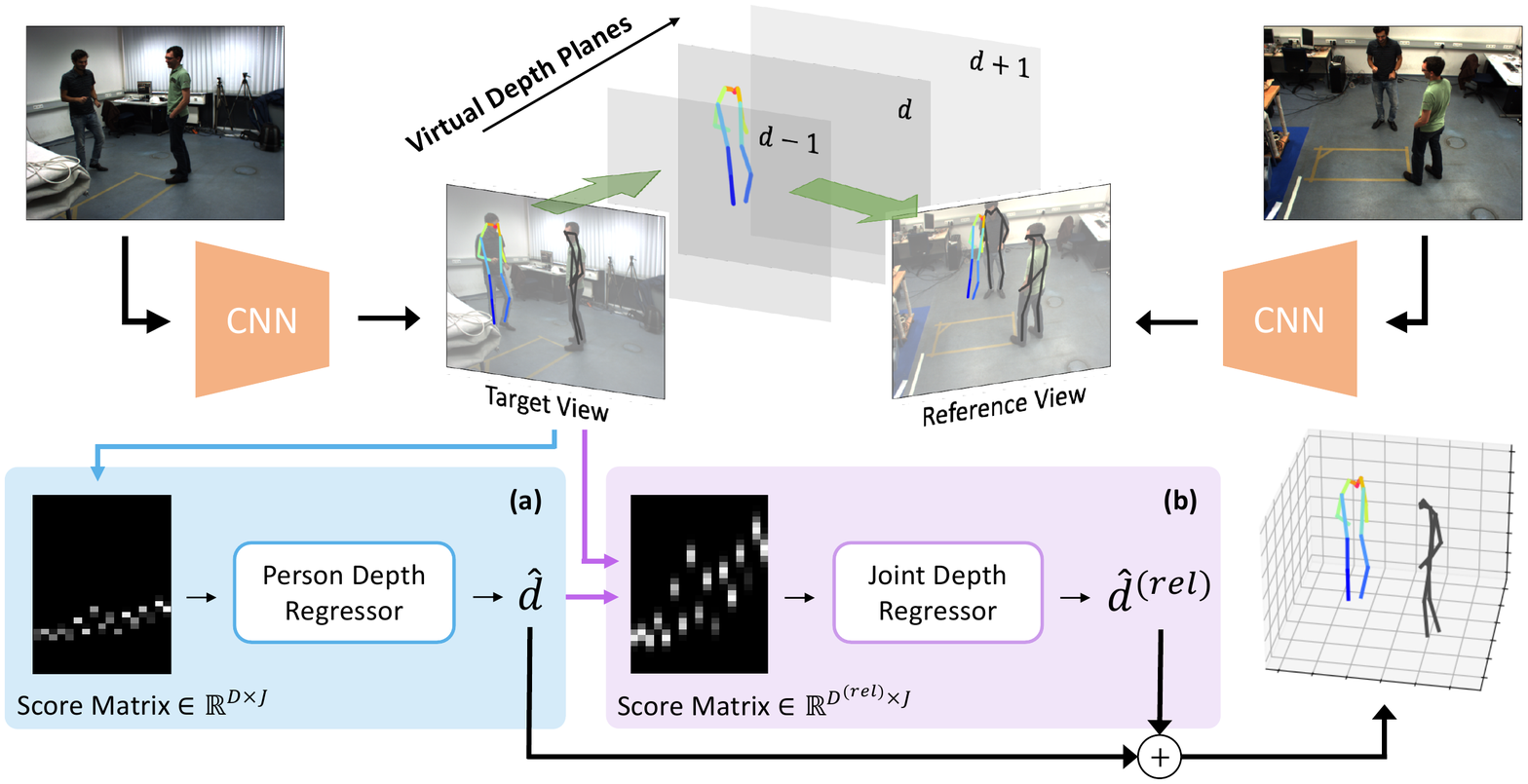}
    \caption{Overview of our approach. 2D pose estimation is first performed for each camera view. We then use the plane sweep algorithm to aggregate the cross-view consistency score for the target person highlighted with jet colormap. Person-level depth is regressed first in (a). Joint-level relative depth is then estimated in (b) and combined with the person-level depth to reconstruct the 3D pose.}
    \label{fig:framework}
\end{figure*}

\subsection{Multi-view Multi-person}

Single-person 3D pose estimation with multi-camera settings has achieved satisfying results on benchmark datasets such as Human3.6M \cite{ionescu2013human3}.
However, it is much more challenging for the multi-person case.
The key difficulty lies in the cross-view matching since the identity of 2D poses from each view is unknown.
Early approaches \cite{belagiannis20143d,belagiannis20153d,ershadi2018multiple} create a common state space shared by all persons via triangulation of corresponding body joints in pairs of the camera views. A 3D pictorial structure is defined as a graphical model with unary and pairwise potentials, and 3D poses are obtained from inference on the graph with the loopy belief propagation algorithm \cite{bishop2006pattern}.
Recent works \cite{kadkhodamohammadi2020generalizable,dong2019fast,chen2020multi,huang2020end} adopt a multi-stage pipeline for the multi-view multi-person 3D pose estimation task.
The pipeline consists of a cross-view matching step to group 2D poses from different views that correspond to the same person, and a 3D pose estimation step to reconstruct the 3D pose from the clustered 2D poses for each person.
Kadkhodamohammadi \textit{et al.} \cite{kadkhodamohammadi2020generalizable} propose to compute a distance between each pair of 2D poses from different views based on the epipolar constraints, and then find the cross-view correspondences with the lowest distance. Instead of directly performing triangulation, the matched 2D poses from all camera views are stacked together and passed into a regression neural network to estimate the 3D pose.
Dong \textit{et al.} \cite{dong2019fast} enhance the cross-view consistency with appearance features. They utilize a person Re-ID network \cite{zhong2018camera} to get the appearance features for each person. These features are then used to compute the appearance-based distance. They also formulate a convex optimization problem to solve for the optimal correspondence matrix, and use a rank constraint to enforce cycle-consistency.
Chen \textit{et al.} \cite{chen2020multi} propose to match cross-view 2D poses by applying the epipolar constraints on feet joints instead of the entire 2D pose. They perform bipartite matching for each pair of views on the pairwise affinities defined on feet joints. A maximum a posteriori (MAP) estimator is adopted for the 3D pose reconstruction.
Huang \textit{et al.} \cite{huang2020end} propose a greedy bottom-up matching approach for 2D pose grouping. Candidate 3D poses are first obtained from triangulation of each pair of 2D poses. These candidate 3D poses form a 3D pose subspace that are then used with a distance-based greedy clustering approach to group the cross-view poses.
Triangulation with learnable weights inspired by \cite{iskakov2019learnable} is applied for each group to obtain the 3D pose estimates.

The aforementioned methods are multi-stage pipelines, where incorrect correspondences can cause large errors in the subsequent 3D pose estimation step.
A recent work, VoxelPose \cite{tu2020voxelpose}, presents a novel pipeline that avoids the explicit cross-view matching and performs 3D pose estimation directly from the multi-view input. This work is inspired by the volumetric approach presented in \cite{iskakov2019learnable} that generates 3D volumes from 2D detections. To identify multiple persons in the common 3D volumetric space, VoxelPose utilizes a 3D object detection formulation to localize each 3D pose, followed by a per-person 3D pose estimation. VoxelPose shows promising results since cross-view consistency is implicitly enforced in the 3D pose estimation. However, the 3D convolution used on the volumetric space is computationally expensive, thus not scalable for larger scenes.

In this work, we present our multi-view 3D pose estimation approach.
Inspired by plane sweep stereo \cite{collins1996space,im2019dpsnet} for dense depth regression, our approach utilizes a pose-aware geometric consistency metric to aggregate multi-view information and performs depth regression for 2D poses without explicitly establishing correspondences.
Our approach demonstrates higher 3D pose estimation precision, while being much more efficient compared to previous works.

\section{Our Method}

Our task is to estimate the 3D poses for all persons in a common 3D space from multi-view images captured by a set of synchronized and calibrated cameras.
The overview of our framework is shown in Figure \ref{fig:framework}.
We first perform 2D pose estimation for each camera view independently using a top-down multi-person pose estimation approach, \textit{e.g.}, HRNet \cite{sun2019deep}.
Subsequently, we perform depth regression for each candidate 2D pose with $J$ joints under a target camera view by utilizing 2D pose detections from multiple reference views.
Finally, the 3D poses can be reconstructed from back-projections of the candidate 2D poses with the estimated depths.
In this section, we present our multi-view depth regression approach based on plane sweep stereo.
A coarse person-level depth regression module is introduced first in Section \ref{subsec:person-level}, followed by a per-person joint-level relative depth regression module in Section \ref{subsec:joint-level}.

\subsection{Person-Level Depth Regression}\label{subsec:person-level}

Our framework is inspired by the plane sweep stereo for dense depth estimation.
The basic idea of plane sweep stereo is to back-project the target view image to a set of successive virtual depth planes, and then warp these projections to the reference view images so that photometric consistency can be measured to determine the depth of each target view pixel.
We adopt the concept of plane sweep in our framework for the person- and joint-level depth regression.
In contrast to the dense depth estimation in the standard plane sweep stereo that relies on photometric consistency, we measure a pose-aware geometric consistency for the depth regression of 2D human poses instead.
In this section, we present a person-level depth regression module to coarsely localize the depth for each candidate 2D pose.
The person-level depth is defined to be the depth of the center hip joint of each person in our implementation.

\subsubsection{Multi-View Score Aggregation}\label{subsec:score_aggregation}

We define a set of $D$ virtual depth planes equally spaced in $[d_{\text{min}}$, $d_{\text{max}}]$ to represent depths in the target camera coordinate frames.  
We set $[d_{\text{min}}$, $d_{\text{max}}]$ such that the depth range is reasonably large to cover the common 3D space shared by multiple cameras.
We empirically set $D=64$ depth planes from experiments (\cf Section \ref{sec:experiments}) in our implementation.

A candidate 2D pose $p$ in the target view is first back-projected to a virtual depth plane $d$, and then followed by a projection to a reference view.
The projected 2D pose is denoted as $q^{(d)}$.
We then search for the nearest 2D pose $r^{(d)}$ from the set of candidate 2D poses $\{p'\}$ in the reference view by:
\begin{equation}
    r^{(d)} = \operatorname*{arg\,min}_{p'} \sum_{j=1}^{J}\tau(p'_j, q_{j}^{(d)}),
\end{equation}
where the function $\tau(\cdot,\cdot)$ measures the distance between two joints in the reference image plane.
Subsequently, we generate a score matrix $\mathcal{S} \in \mathbb{R}^{D \times J}$ for the target pose $p$.
The score of joint $j$ at depth $d$ measures the alignment of the projected pose $q^{(d)}$ with the matched reference view pose $r^{(d)}$ at joint $j$. It is computed as:
\begin{equation}
    \mathcal{S}_{d, j} = \exp{\Bigg\{-\frac{{\big[\tau(r_{j}^{(d)},q_{j}^{(d)})\big]}^2}{2 \cdot {\sigma}^2}\Bigg\}}.
    \label{eqn:score}
\end{equation}
A small distance between joint $r_{j}^{(d)}$ and joint $q_{j}^{(d)}$ results in a high score of $\mathcal{S}_{d,j}$. This indicates a higher chance for the depth of joint $p_j$ to be around $d$.
$\sigma$ is a hyper-parameter to control the width of the bell curve.

Figure \ref{fig:framework}(a) shows an example of the score matrix.
It is a measurement of the pose-aware cross-view geometric consistency and is used for the subsequent depth regression.
In cases when multiple reference views are available, we fuse the score matrices computed from all reference views via a weighted averaging, where the confidence of the matched 2D pose $r^{(d)}$ in each reference view obtained from the 2D pose estimator is used as the weight.

\vspace{-3mm} \paragraph{Remark:}
Note that back-projecting all joints of a 2D pose to the same depth plane and projecting the ``flat" 3D pose to the reference view for pose matching is an approximation. However, it does not affect the retrieval of the nearest pose from the reference view in most scenarios and works sufficiently well in practice.

\begin{figure*}[t]
    \centering
    \includegraphics[width=\linewidth, trim=30 180 80 100, clip]{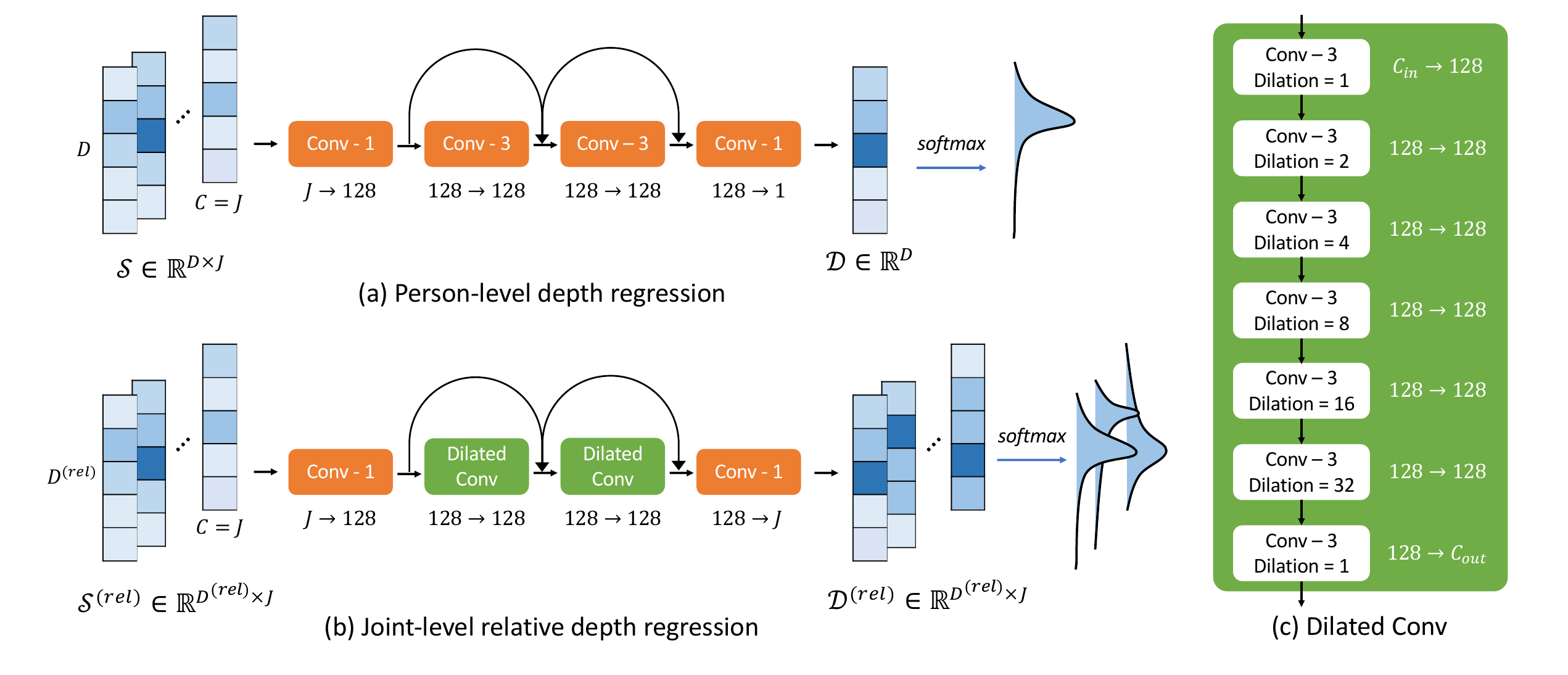}
    \caption{Network structure of (a) the person-level depth regression network, (b) the joint-level depth regression network, and (c) the dilated convolution module used in (b). Each block ``Conv - $k$'' consists of a 1D convolution with kernel size $k$ followed by the batch normalization \cite{ioffe2015batch} and ReLU operations. The numbers in $A \rightarrow B$ denote the channel size. Residual links are used in both networks.}
    \vspace{-2mm}
    \label{fig:network}
\end{figure*}

\subsubsection{Depth Regression}

We treat the score matrix $\mathcal{S} \in \mathbb{R}^{D \times J}$ 
of a target pose $p$
as a 1D-signal of length $D$ with $J$ feature channels, and
utilize a 1D Convolutional Neural Network (1D-CNN) to map it into a depth vector $\mathcal{D} \in \mathbb{R}^{D}$.
As illustrated in Figure \ref{fig:network}(a), 
we use a simple architecture with residual links that is sufficient to coarsely estimate the person-level depth.

A soft-argmax operation can be applied on the output depth vector $\mathcal{D}$ to obtain the scalar depth value $\hat{d}$:
\begin{equation}
    \hat{d} = \sum_{i=1}^{D} d_i \cdot \mathcal{D}_i,
\end{equation}
where $d_i$ is the depth of the $i^\text{th}$ depth layer.
Despite its effectiveness on the single-person case \cite{sun2018integral}, soft-argmax operation assumes uni-modality of the input distribution, which can fail in multi-person scenarios.
To overcome this limitation, we propose to use an adapted ``local'' soft-argmax instead:
\begin{equation}
    \hat{d} = \frac{\sum_{i=i'}^{i'+\delta -1} d_i \cdot \mathcal{D}_i}{\sum_{i=i'}^{i'+\delta -1} \mathcal{D}_i}, ~\text{where}~ i' = \operatorname*{arg\,max}_{i''} \sum_{i=i''}^{i''+\delta -1} \mathcal{D}_i.
    \label{eqn:soft-argmax}
\end{equation}
A window of size $\delta$ slides over $\mathcal{D}$ to search for the window with the largest response from $\mathcal{D}$.
Standard soft-argmax is then computed within the window to obtain $\hat{d}$.
We use a window size $\delta = D/4 = 16$ in our implementation.
Note that when the window size is equal to the length of the input signal, \textit{i.e.}, $\delta = D$, Equation \ref{eqn:soft-argmax} degenerates to the standard soft-argmax operation.

The network is trained by minimizing the $L_1$ loss between the regressed depth $\hat{d}$ and the ground truth person-level depth $d_{*}$ over all 2D poses $\{ p\}$ in the target view:
\begin{equation}
    \mathcal{L}_{\text{pose}} = \sum_{p}||\hat{d}(p) - d_{*}(p)||_1.
\end{equation}

\vspace{-3mm} \paragraph{Remark:}
Although we aggregate the scores at each depth layer independently, the scores at successive depth layers exhibit smooth variation (see the score matrix in Figure \ref{fig:framework}(a) for an illustration).
1D-CNN across the depth dimension effectively aggregates local features over all $J$ joints at successive depth layers, facilitating the regression of the coarse person-level depth.

\subsection{Joint-Level Relative Depth Regression}\label{subsec:joint-level}

\subsubsection{Score Aggregation}
After the coarse localization of each 2D pose with the regressed person-level depth,
we adopt a fine-grained joint-level relative depth regression module to estimate the per-joint relative depth with respect to the person-level depth.
Similar to the person-level depth regression module in Section \ref{subsec:score_aggregation}, a score matrix $\mathcal{S}^{(rel)}$ is aggregated from the reference views for each joint $j$ at each relative depth layer $d^{(rel)}$:
\begin{equation}
    \mathcal{S}_{d^{(rel)}, j}^{(rel)} = \exp{\Bigg\{-\frac{{\big[\tau(r_{j}^{(\hat{d} + d^{(rel)})},q_{j}^{(\hat{d} + d^{(rel)})})\big]}^2}{2 \cdot {\sigma}^2}\Bigg\}}.
    \label{eqn:score_rel}
\end{equation}
The key difference is that we use a different set of $D^{(rel)}$ virtual depth planes in the range of $[-1000, +1000]$mm, which is sufficient to cover the depth range of arbitrary pose variation. $D^{(rel)}$ is also set to 64 in our implementation.
Note that $\hat{d}$ is the estimated person-level depth from Equation \ref{eqn:soft-argmax}.
During training, we use the ground truth person-level depth $d_{*}$ in place of $\hat{d}$ in Equation \ref{eqn:score_rel} to stabilize the training process.

\subsubsection{Depth Regression}

We use another 1D-CNN to regress the per-joint relative depth from the score matrix $\mathcal{S}^{(rel)}$.
The network structure is shown in Figure \ref{fig:network}(b).
Since the joint-level depth planes compactly surround each target person, it is expected to see more widely-spread peaks in the score matrix (See Figure \ref{fig:framework}(b) for an illustration of the score matrix).
Consequently, compared to the person-level depth regression, a larger receptive field along the depth dimension is needed for the joint-reasoning of the depths for all body joints.
To this end, we use a series of 1D dilated convolutions to effectively increase the receptive field as illustrated in Figure \ref{fig:network}(c).
The output of the network is the relative depth matrix $\mathcal{D}^{(rel)} \in \mathbb{R}^{D^{(rel)} \times J}$.
The relative depth of each joint is obtained by the standard soft-argmax operation:
\begin{equation}
    \hat{d}_{j}^{(rel)} = \sum_{i=1}^{D^{(rel)}}d_{i}^{(rel)} \cdot \mathcal{D}_{i,j}^{(rel)},
\end{equation}
where $d_{i}^{(rel)}$ is the relative depth of the $i^\text{th}$ depth layer.

Similarly, the $L_1$ loss between the regressed relative depth $\hat{d}^{(rel)}$ and the ground truth relative depth $d_{*}^{(rel)}$ is minimized:
\begin{equation}
    \mathcal{L}_{\text{joint}} = \sum_p \sum_j ||\hat{d}(p)_{j}^{(rel)} - d_{*}(p)_{j}^{(rel)}||_1.
\end{equation}
During inference, the absolute depth of each joint is computed by:
\begin{equation}
    \hat{d}_{j}^{(abs)} = \hat{d} + \hat{d}_{j}^{(rel)},
\end{equation}
which is then used to back-project the 2D pose to produce the final 3D pose estimate.

\vspace{-3mm} \paragraph{Remark:}
Our framework adopts a coarse-to-fine scheme to decouple the task into a person-level depth regression and a per-person joint-level relative depth regression.
The benefit of utilizing a two-stage scheme is that we can reduce the computational cost by using a sparse set of virtual depth layers in the first stage for a coarse depth regression from a larger depth range.
Since the joint-level relative depth regression is able to compensate for small person-level depth offsets, the person-level depth in the first stage does not need to be very precise 
to achieve an accurate final 3D pose estimation.
The two-stage practice is also widely used in object detection pipelines such as \cite{ren2015faster,he2017mask}.

\subsection{Training Details}

In view of the limited availability of multi-view multi-person 3D pose annotations, we follow the practice in \cite{tu2020voxelpose} to use synthesized data in training both the person-level and joint-level depth regression modules.
Specifically, we utilize 3D pose skeletons from a MoCap dataset and randomly place them in a pre-defined 3D space.
The 3D poses are projected to 2D poses under each camera view which serve as the input to our depth regression modules.
We randomly perturb the image coordinates of 2D poses in different views to simulate the scenario of in-precise 2D pose estimation.
Confidence score is assigned to each 2D joint based on the level of random perturbation.

During inference, we use the 2D pose estimator HRNet \cite{sun2019deep} to obtain candidate 2D poses.
We take each camera view as the target view in turn to generate 3D pose estimates given the depth estimation under that particular view.
3D poses from all camera views are fused into the same global coordinate space.
Duplicates can be effectively removed by clustering and averaging nearby 3D poses given a distance threshold.

\begin{table}[t]
    \centering
    \resizebox{\columnwidth}{!}{
        \begin{tabular}{c|cccc}
            \toprule
            Campus & Actor 1 & Actor 2 & Actor 3 & Average \\
            \midrule
            standard soft-argmax  & 98.0 & 93.2 & 97.7 & 96.3 \\
            ``local'' soft-argmax & \textbf{98.4} & \textbf{93.7} & \textbf{99.0} & \textbf{97.0} \\
            \midrule
            \midrule
            Shelf & Actor 1 & Actor 2 & Actor 3 & Average \\
            \midrule
            standard soft-argmax  & 99.1 & 95.7 & 98.0 & 97.6 \\
            ``local'' soft-argmax & \textbf{99.3} & \textbf{96.5} & \textbf{98.0} & \textbf{97.9} \\
            \bottomrule
        \end{tabular}
    }
    \caption{Comparison of PCP between the soft-argmax operations used in the person-level depth regression on the Campus and the Shelf datasets.}
    \label{tab:soft-argmax}
\end{table}

\section{Experiments}\label{sec:experiments}

\subsection{Datasets and Metrics}

\paragraph{Campus \cite{belagiannis20143d}.}
The Campus dataset captures an outdoor environment with three persons interacting with each other using three cameras.
Due to the incomplete annotation of 3D ground truth poses, we directly use HRNet \cite{sun2019deep} pre-trained on COCO \cite{lin2014microsoft} to estimate the 2D poses and train our depth regression modules with synthesized 3D MoCap poses.
We follow previous works \cite{dong2019fast,tu2020voxelpose,huang2020end} and perform evaluation on the test set frames: 350-470, 650-750.

\vspace{-3mm} \paragraph{Shelf \cite{belagiannis20143d}.}
The Shelf dataset captures an indoor environment with four persons interacting with each other using five cameras.
Similar to the Campus dataset, we use pre-trained HRNet to estimate 2D poses and only train our depth regression modules with synthesized data.
We follow previous works \cite{dong2019fast,tu2020voxelpose,huang2020end} in evaluating only three of the four persons on the test set frames: 300-600 since one person is occluded in majority of the frames.

\vspace{-3mm}\paragraph{CMU Panoptic \cite{joo2017panoptic}}
The dataset captures an indoor environment with multiple actors performing social activities.
Following \cite{tu2020voxelpose}, we use HRNet pre-trained on COCO and fine-tuned on Panoptic to obtain 2D poses.
We use the same set of training and testing sequences captured by the same set of HD cameras (3, 6, 12, 13, 23) as in \cite{tu2020voxelpose} for evaluation.

\vspace{-3mm} \paragraph{Evaluation metrics.}
Following \cite{dong2019fast,huang2020end,tu2020voxelpose}, we use the Percentage of Correctly estimated Parts (PCP) to evaluate the accuracy of the estimated 3D poses for the Campus and the Shelf datasets.
Specifically, the closest estimated 3D pose is selected to evaluate the correctness of each body part for each ground truth pose.
To better understand the performance of the person-level depth regression module, we also evaluate the recall rate of the person-level depth at various error thresholds.
Since previous works share no common evaluation protocol on the Panoptic dataset, we follow the evaluation process in VoxelPose \cite{tu2020voxelpose} and report the Average Precision (AP) and Mean Per Joint Position Error (MPJPE).

\vspace{+3mm}
\subsection{Ablation Study}

\paragraph{``Local'' Soft-argmax.}
We first justify the use of a ``local'' soft-argmax operation (\cf Equation \ref{eqn:soft-argmax}).
A direct comparison between using a standard soft-argmax and our ``local'' soft-argmax is shown in Table \ref{tab:soft-argmax}.
The ``local'' soft-argmax is able to focus on the mode with the highest response without being interfered by other modes in a multi-modality distribution.
From Table \ref{tab:soft-argmax}, we see obvious improvement when using our ``local'' soft-argmax operation on both the Campus and the Shelf datasets.
The performance gap is large especially for the Campus dataset, where each camera view typically consists of 2 to 3 persons.

\begin{table}[t]
    \centering
    \resizebox{0.8\columnwidth}{!}{
        \begin{tabular}{lccccc}
            \toprule
            & \# Cameras & $D$ & $D^{(rel)}$ & PCP (\%) \\
            \midrule
            \midrule
            (a)    & 5 & 64 & 64 & 97.9 \\
            (b)    & 5 & 16 & 64 & 97.5 \\
            (c)    & 5 & 64 & 16 & 97.2 \\
            \midrule
            (d)    & 4 & 64 & 64 & 97.5 \\
            (e)    & 3 & 64 & 64 & 95.1 \\
            (f)    & 2 & 64 & 64 & 92.5 \\
            \midrule
            (g)$+$ & 5 & 64 & 64 & 97.7 \\
            \bottomrule
        \end{tabular}
    }
    \caption{Ablation study on the Shelf dataset. $D$ and $D^{(rel)}$ are the number of depth layers in the person-level and joint-level depth regression modules, respectively. $+$ means different sets of cameras are used for training and evaluation.}
    \label{tab:ablation}
\end{table}

\begin{figure}[t]
    \centering
    \includegraphics[width=\columnwidth]{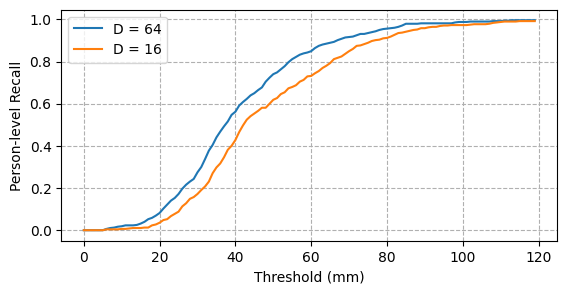}
    \caption{Comparison of person-level recall when different number of depth planes is used.}
    \label{fig:recall}
    \vspace{-2mm}
\end{figure}

\vspace{-3mm} \paragraph{}
We then conduct ablation studies to evaluate our approach under various settings.
The performance of 3D pose estimation on the Shelf dataset measured in PCP is reported in Table \ref{tab:ablation}.

\vspace{-3mm} \paragraph{Number of virtual depth planes.}
We use $D = 64$ person-level depth planes and $D^{(rel)} = 64$ joint-level depth planes in our implementation by default.
In this ablation study, we examine the results when fewer depth planes are used in each stage.

We first compare the person-level depth regression performance by evaluating the recall rate of the center hip joint with respect to various distance thresholds.
The results of $D = 16$ and $64$ are shown in Figure \ref{fig:recall}.
Using more virtual depth planes in general increases the person-level depth estimation precision.
Table \ref{tab:ablation}(a) and (b) also show that the 3D pose estimation accuracy drops by 0.4\% when reducing $D$ from 64 to 16.

The impact from using fewer depth layers is small in the person-level stage since regressing a coarse person-level depth is sufficient for rough 3D localization.
In comparison, the joint-level stage requires more precise depth regression in order to estimate the 3D pose accurately.
Table \ref{tab:ablation}(a) and (c) show the comparison between using 16 and 64 joint-level depth planes.
Reducing $D^{(rel)}$ from 64 to 16 leads to a larger 0.7\% performance drop, which is due to that using fewer depth planes increases the quantization error.

\begin{table}[t]
    \centering
    \begin{tabular}{r|cc}
        \toprule
        Method & Campus (fps) & Shelf (fps) \\
        \midrule
        \midrule
        VoxelPose \cite{tu2020voxelpose} & 5.5 & 3.0 \\
        Ours & 110.0 & 42.8 \\
        \bottomrule
    \end{tabular}
    \caption{Comparison of inference speed (frames per second) with \cite{tu2020voxelpose} on the Campus and the Shelf datasets.}
    \label{tab:running_time}
    \vspace{-2mm}
\end{table}

\vspace{-3mm} \paragraph{Number of cameras.}
In Table \ref{tab:ablation}(a) and (d)-(f), we compare the performance of our approach when different number of cameras is used.
The performance drops with reducing number of cameras.
This is as expected since the indoor dataset exhibits severe occlusions, which can be ambiguous even for multi-camera settings.
Nonetheless, our approach still achieves over 92\% accuracy in the extreme cases of using only two cameras.

\vspace{-3mm} \paragraph{Generalization to different camera settings.}
In this setting, we use randomly sampled camera viewpoints during training and use the 5 cameras from the dataset for evaluation.
We can see from the result in Table \ref{tab:ablation}(g) that our method is able to perform equally well compared to (a) when trained and evaluated on different sets of cameras.
This demonstrates the generalization ability of our method.

\renewcommand{\arraystretch}{1}
\setlength{\tabcolsep}{8pt}
\begin{table*}[t]
    \centering
    \begin{tabular}{r|cccc|cccc}
        \toprule
        & \multicolumn{4}{c|}{Campus} & \multicolumn{4}{c}{Shelf} \\
        Method & Actor 1 & Actor 2 & Actor 3 & Average & Actor 1 & Actor 2 & Actor 3 & Average\\
        \midrule
        \midrule
        Belagiannis \textit{et al.} \cite{belagiannis20143d} & 82.0 & 72.4 & 73.7 & 75.8 & 66.1 & 65.0 & 83.2 & 71.4 \\
        Belagiannis \textit{et al.} \cite{belagiannis2014multiple} & 83.0 & 73.0 & 78.0 & 78.0 & 75.0 & 67.0 & 86.0 & 76.0 \\
        Belagiannis \textit{et al.} \cite{belagiannis20153d} & 93.5 & 75.7 & 84.4 & 84.5 & 75.3 & 69.7 & 87.6 & 77.5 \\
        Ershadi-Nasab \textit{et al.} \cite{ershadi2018multiple} & 94.2 & 92.9 & 84.6 & 90.6 & 93.3 & 75.9 & 94.8 & 88.0 \\
        Dong \textit{et al.} \cite{dong2019fast} & 97.6 & 93.3 & 98.0 & 96.3 & 98.8 & 94.1 & 97.8 & 96.9 \\
        Huang \textit{et al.} \cite{huang2020end} & 98.0 & \textbf{94.8} & 97.4 & 96.7 & 98.8 & 96.2 & 97.2 & 97.4 \\
        VoxelPose - Tu \textit{et al.} \cite{tu2020voxelpose} & 97.6 & 93.8 & 98.8 & 96.7 & \textbf{99.3} & 94.1 & 97.6 & 97.0 \\
        Ours & \textbf{98.4} & 93.7 & \textbf{99.0} & \textbf{97.0} & \textbf{99.3} & \textbf{96.5} & \textbf{98.0} & \textbf{97.9} \\
        \bottomrule
    \end{tabular}
    \vspace{+1mm}
    \caption{Comparison of PCP with existing multi-view multi-person 3D pose estimation methods on the Campus and the Shelf datasets.}
    \vspace{-4mm}
    \label{tab:benchmark}
\end{table*}

\begin{figure*}
    \centering
    \includegraphics[width=\linewidth, trim=20 210 60 0, clip]{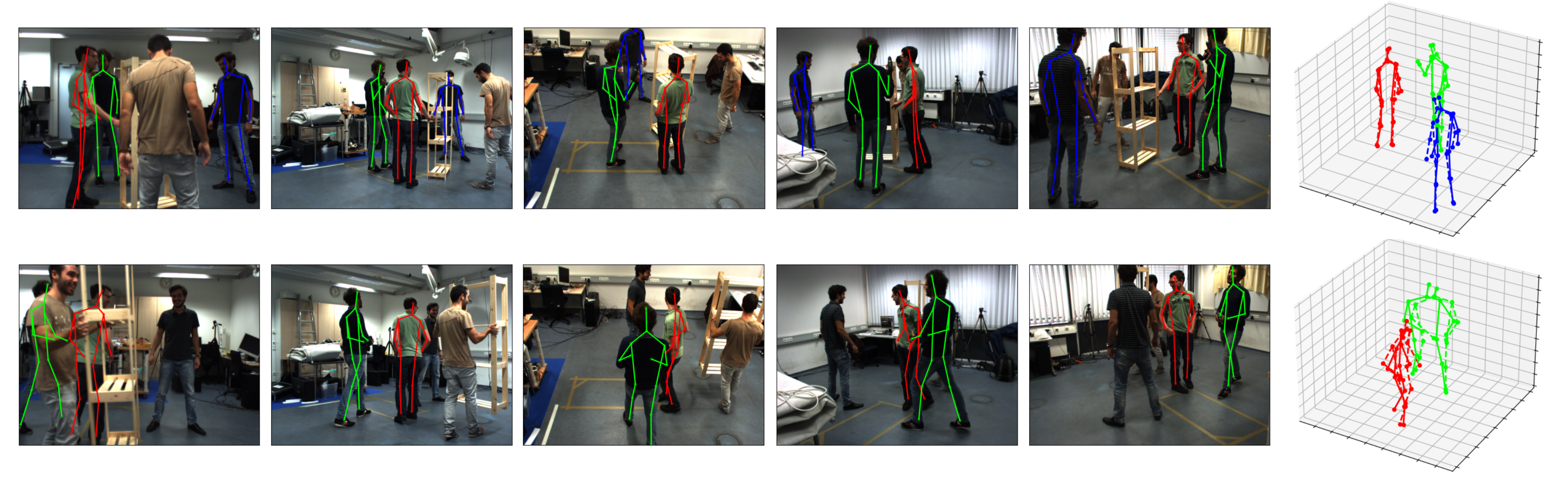}
    \vspace{-6mm}
    \caption{Qualitative results on the Shelf dataset. (Left) Images with projections of the estimated 3D poses in each camera view. (Right) 3D pose estimates in solid lines and 3D ground truth poses in dashed lines.}
    \label{fig:qualitative}
    \vspace{-1mm}
\end{figure*}

\renewcommand{\arraystretch}{1}
\setlength{\tabcolsep}{6pt}
\begin{table}
    \centering
    \resizebox{\columnwidth}{!}{
        \begin{tabular}{r|ccccc}
            \toprule
            Method & $\text{AP}_{25}$ & $\text{AP}_{50}$ & $\text{AP}_{100}$ & $\text{AP}_{150}$ & MPJPE \\
            \midrule
            \midrule
            VoxelPose \cite{tu2020voxelpose} & 83.59 & 98.33 & 99.76 & \textbf{99.91} & 17.68mm \\
            Ours      & \textbf{92.12} & \textbf{98.96} & \textbf{99.81} & 99.84 & \textbf{16.75mm} \\
            \bottomrule
        \end{tabular}
    }
    \caption{Comparison with \cite{tu2020voxelpose} on CMU Panoptic dataset.}
    \label{tab:panoptic}
    \vspace{-3mm}
\end{table}

\subsection{Inference Speed}

We compare the computational efficiency of our work with VoxelPose \cite{tu2020voxelpose} since both works address the multi-view 3D pose estimation problem in a learning-based framework without explicit cross-view matching.
We perform inference for both methods using the same set of 2D pose estimates on a single GTX 1080 Ti graphics card.
The inference runtime for the evaluation on both the Campus and the Shelf datasets is reported in Table \ref{tab:running_time}.
Note that the runtime for 2D pose estimation is not included.
Our method achieves a frame rate that is up to 20x faster than the VoxelPose method, which can be used to fully support real-time inference.
Our advantage in computational efficiency is mainly due to our method focusing on only the target 2D poses and uses more computationally affordable 1D convolutions.
In contrast, VoxelPose performs expensive 3D convolutions for each voxel in the 3D space that incurs unnecessary computations in regions without any person.

\subsection{Comparison to the State-of-the-arts}

We report the 3D pose estimation accuracy on the Campus and the Shelf datasets in Table \ref{tab:benchmark}.
The average accuracy improves from 96.7\% to 97.0\% on the Campus dataset, and from 97.4\% to 97.9\% on the Shelf dataset.
Table \ref{tab:panoptic} further shows that our method decreases the error by $\sim$1mm on Panoptic dataset when compared to VoxelPose \cite{tu2020voxelpose}.
Our method shows decent performance improvement in addition to being able to implicitly solve the challenging multi-view matching problem neatly.
Note that among the existing works shown in Table \ref{tab:benchmark}, VoxelPose \cite{tu2020voxelpose} and our method utilize only geometric consistency based on multi-view epipolar geometry.
Although photometric consistency is not considered, robust performance can still be achieved with 2D poses from the current top-performing 2D pose estimator.
Examples of qualitative results are shown in Figure \ref{fig:qualitative}.

\section{Conclusion}

In this work, we present our plane-sweep-based approach to regress 2D pose depths for the task of multi-view multi-person 3D pose estimation.
Our method uses the plane sweep algorithm to aggregate multi-view information based on a pose-aware geometric consistency and effectively estimates the depths for each 2D pose in a target camera view without explicitly establishing cross-view correspondences.
Depth regression is performed in a coarse-to-fine scheme, where we first regress the person-level depth followed by the joint-level relative depth estimation.
Our framework is computationally more efficient and shows superior performance compared to previous state-of-the-arts.


\clearpage

{\small
\bibliographystyle{ieee_fullname}
\bibliography{submission}
}

\end{document}